\documentclass[10pt,twocolumn]{article} 
\usepackage{simpleConference}
\usepackage{times}
\usepackage{graphicx}
\usepackage{amssymb}
\usepackage{url,hyperref}
\usepackage{booktabs}
\usepackage{amsmath}

\usepackage[linesnumbered,ruled,lined]{algorithm2e}
\usepackage{algpseudocode}
\usepackage{multirow}

\SetKwFor{For}{for}{do}{endfor} 
\newtheorem{definition}{Definition}

\begin{document}
	
\title{HMSG: Heterogeneous Graph Neural Network based on  Metapath Subgraph Learning}

\author{Xinjun Cai$^{1,2}$, Jiaxing Shang$^{1,2,*}$, Fei Hao$^{3}$, Dajiang Liu$^{1,2}$, Linjiang Zheng$^{1,2}$\\
	\\
	$^{1}$ College of Computer Science, Chongqing University, Chongqing 400044, China \\
	$^{2}$ Key Laboratory of Dependable Service Computing in Cyber Physical Society, \\Ministry of Education, Chongqing University, Chongqing 400044, China \\
	$^{3}$ School of Computer Science, Shaanxi Normal University, Shaanxi 710062, China \\
	$^{*}$ Corresponding author: Jiaxing Shang, Email: shangjx@cqu.edu.cn \\
}

\maketitle
\thispagestyle{empty}

\begin{abstract}
	Many real-world data can be represented as heterogeneous graphs with different types of nodes and connections. Heterogeneous graph neural network model aims to embed nodes or subgraphs into low-dimensional vector space for various downstream tasks such as node classification, link prediction, etc. Although several models were proposed recently, they either only aggregate information from the same type of neighbors, or just indiscriminately treat homogeneous and heterogeneous neighbors in the same way. Based on these observations, we propose a new heterogeneous graph neural network model named HMSG to comprehensively capture structural, semantic and attribute information from both homogeneous and heterogeneous neighbors. Specifically, we first decompose the heterogeneous graph into multiple metapath-based homogeneous and heterogeneous subgraphs, and each subgraph associates specific semantic and structural information. Then message aggregation methods are applied to each subgraph independently, so that information can be learned in a more targeted and efficient manner. Through a type-specific attribute transformation, node attributes can also be transferred among different types of nodes. Finally, we fuse information from subgraphs together to get the complete representation. Extensive experiments on several datasets for node classification, node clustering and link prediction tasks show that HMSG achieves the best performance in all evaluation metrics than state-of-the-art baselines.
\end{abstract}

\textbf{Keywords:} Heterogeneous graph, Graph neural network, Graph embedding, Metapath

\section{Introduction}
Many real-world data can be organized as graph or network structure, which provides a more abstract representation of various objects and their interactions, such as social networks, traffic networks, protein molecular structures, recommendation systems, etc. Although there are many widely used deep learning models on Euclidean data such as CNN and RNN, they cannot be directly transferred to non-Euclidean data such as graphs due to the irregular structure \cite{henaff2015deep}. Therefore, it is necessary to design deep learning models suitable for graph data.

In the past decade, a large number of graph representation learning methods have been proposed. Random-walk-based methods such as DeepWalk \cite{perozzi2014deepwalk} and node2vec \cite{grover2016node2vec} use random walk to sample node sequences on the graph, and then feed them to the skip-gram \cite{mikolov2013efficient} model to obtain low-dimensional vector representation of each node. Model \cite{scarselli2008graph} and \cite{li2015gated} utilized recurrent neural network to tackle graph data. Due to the powerful feature extraction capabilities of convolutional neural networks, many graph convolutional neural networks have achieved great achievements. ChebNet \cite{defferrard2016convolutional} and GCN \cite{kipf2016semi} use Fourier transform to map graph data to spectral domain for convolution operation. Models such as GraphSAGE \cite{hamilton2017inductive} and GAT \cite{velivckovic2017graph} directly aggregate information from neighbors, which exhibit good generalization ability and stability.

Although graph embedding methods have achieved good performance, most of them deal with homogeneous graphs, in which all nodes and edges in the graph are of the same type. However, in reality, the types of nodes or edges in graphs are usually heterogeneous, such as authors and papers in a scholar graph, or users and items in recommendation systems, etc. These graphs are generally called heterogeneous information networks (HIN) or heterogeneous graphs, which contain abundant structural and semantic information. Recently, scholars have devoted great research efforts on heterogeneous graphs representation learning.

\begin{figure}[h]
	\centering
	\includegraphics[width=\linewidth]{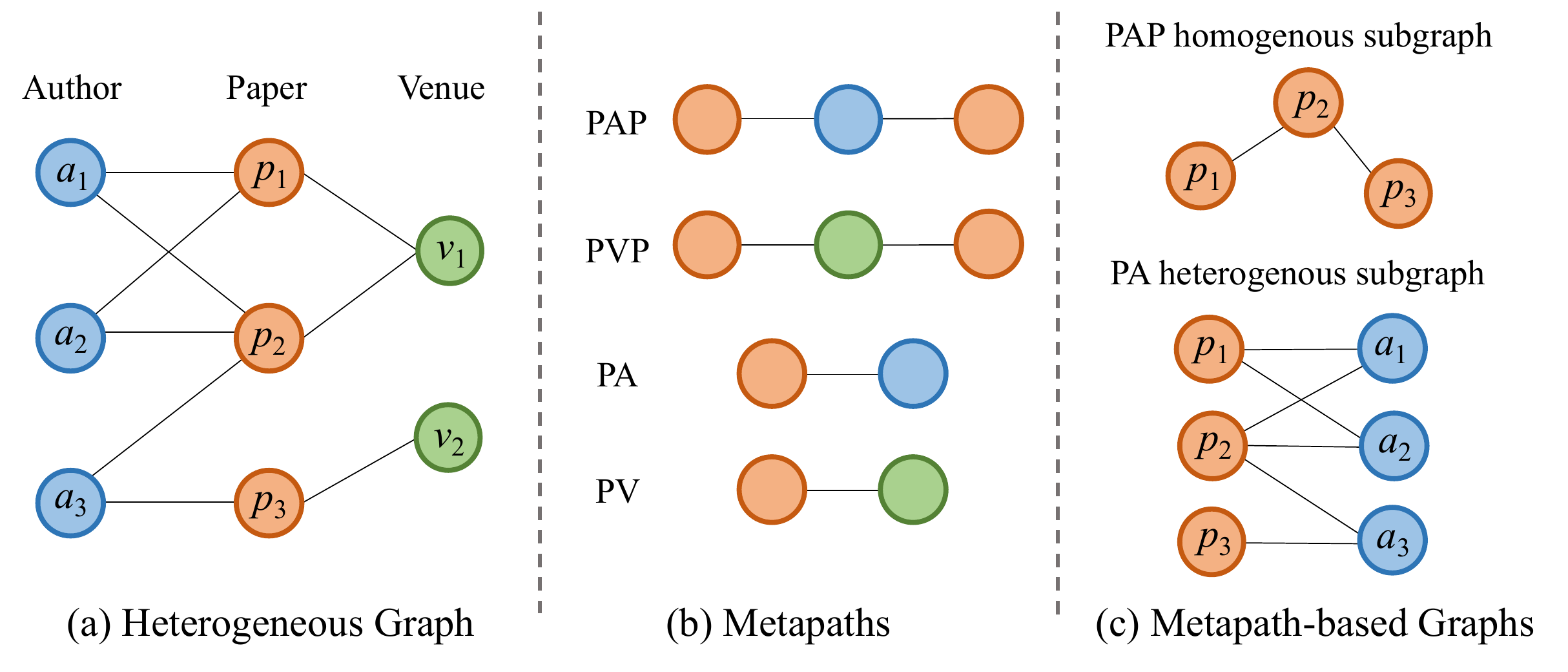}
	\caption{(a) An example heterogeneous graph with three types of nodes (i.e., authors, papers, venues). (b) Four examples of metapath: Paper-Author-Paper (PAP), Paper-Venue-Paper (PVP), Paper-Author (PA) and Paper-Venue (PV). (c) The metapath-based homogeneous graph and heterogeneous  graph, respectively.}
	\label{Fig1}
\end{figure}

\textit{Motivation.} Most of the researches on heterogeneous graphs are based on metapath \cite{sun2012mining}. A metapath is an ordered sequence of node types and edge types, which is used to capture specific semantic information of the graph. For example, a simple co-author graph is shown in Figure \ref{Fig1} (a), which contains three types of nodes: author, paper, and venue. The metapaths such as paper-author-paper (PAP) and paper-venue-paper (PVP) represent relationships between two papers, paper-author (PA) and paper-venue (PV) represent interaction relation between papers and authors or venues. Based on this, methods such as metapath2vec \cite{dong2017metapath2vec}, HIN2vec \cite{dong2017metapath2vec} and HERec \cite{shi2018heterogeneous} embed heterogeneous graph structure into low-dimensional vectors. HetGNN \cite{zhang2019heterogeneous}, HAN \cite{wang2019heterogeneous} and MAGNN \cite{fu2020magnn} further incorporate attributes of nodes and aggregate embeddings of multiple metapaths or neighbor sets with attention mechanism. However, these heterogeneous graph models mainly suffer from the following two limitations: First, many of them only aggregate information from homogeneous neighbors connected by metapaths, and discard rich structural and attribute information of heterogeneous neighbors; Second, some studies aggregate information from both homogeneous and heterogeneous neighbors, but just indiscriminately treat these neighbors in the same way. As a consequence, these methods may lose important information and result in unsatisfactory performance.

We show this through the following example. As shown in the scholar graph in Figure \ref{Fig1} (c), through metapath PAP a homogeneous subgraph with paper nodes $p_1,p_2$ and $p_3$ is constructed. For a specific paper node (e.g., $p_2$), if we only aggregate information from the homogeneous neighbors ($p_1$ and $p_3$), then the structural and attribute information of heterogeneous neighbors ($a_1,a_2,a_3$) that contribute to their connections will be ignored. Therefore, only considering homogeneous graphs will lose a lot of useful interactive information in the original graph, and these interaction relationships are especially important to some tasks such as link prediction. Moreover, there are different interactive relationships between a node and different types of neighbors. Since these relationships often carry different semantics, they should be considered separately to avoid information loss. It is worth mentioning that the attributes of different types of nodes are often different. For example, in the recommendation systems, the attributes of user nodes are generally age, gender, hobbies, etc., while items have attributes like price, text description, and images. Raw attributes cannot be directly transferred between different types of nodes, and need to be transformed in advance. 

To the best of our knowledge, no existing studies have simultaneously considered all the above aspects. Based on this observation, in this paper we propose a new heterogeneous graph representation learning model named HMSG (\textbf{H}eterogeneous graph neural network based on  \textbf{M}etapath \textbf{S}ub\textbf{G}raph learning) which comprehensively captures structural, semantic and attribute information from both homogeneous and heterogeneous neighbors. To this end, we first perform a type-specific attribute transformation to project the attributes of different types of nodes into the same latent space, so that attribute information can be transferred among them. Then, in order to learn more discriminatively, we generate multiple homogeneous and heterogeneous subgraphs\footnote{Note the concept of subgraph in this paper is different from that in traditional graph theory.} from the original heterogeneous graph through various metapaths. This step can be regarded as task decomposition, because after being decomposed into subgraphs, the originally complex structural and semantic information can be learned in a more targeted and efficient manner. By learning from both homogeneous and heterogeneous subgraphs independently, HMSG not only aggregates information from homogeneous neighbors, but also obtain attribute and structural information from heterogeneous neighbors. On each subgraph, different aggregation methods can be applied to get the node representation. Finally, we perform attention-based aggregation to combine information from different subgraphs according to their importance, resulting the final node representation.

In summary, the main contributions of our work are as follows:
\begin{itemize}
	\item We propose HMSG, a metapath-based heterogeneous graph neural network model which comprehensively captures structural, semantic and attribute information from both homogeneous and heterogeneous neighbors.
	
	\item We creatively decompose the heterogeneous graph representation learning task into multiple metapath-based subgraph learning tasks, so that the originally complex structural and semantic information can be learned in a more targeted and efficient manner.
	
	\item Experiments of node classification, node clustering and link prediction are carried out on multiple datasets, and our proposed model achieves the best performance in all metrics than other state-of-the-art baselines.
\end{itemize}

\section{Related Work}

\subsection{Graph Neural Networks}
The purpose of GNN is to apply deep neural network model to graph representation learning, and map graphs to low-dimensional vector spaces for downstream tasks, such as link prediction \cite{zhang2018link,chen2018gc}, node classification \cite{rong2019dropedge, abu2020n}, recommendation \cite{qiu2020exploiting, zhao2017meta, fan2019graph}, etc. The notion of graph neural networks is initially outlined in \cite{gori2005new}, then Franco et al. \cite{scarselli2008graph} extended recursive neural networks to graph learning task and Li et al. \cite{li2015gated} treated the neighborhood information as the time step input of gated recurrent units. Recently, how to apply convolutional neural networks to graph data has become a hot topic. Related works can be divided into two categories: spectral-based methods and spatial-based methods. The main idea of spectral-based methods is to perform convolution operations in the Fourier domain. Defferrard et al. \cite{defferrard2016convolutional} used Chebyshev polynomials to approximate the graph filters, which reduced the complexity of graph convolution in spectral domain. Kipf et al. \cite{kipf2016semi} further restricted the filters only on the first-order neighborhood of each node. Due to the reason that spectral-based methods input the whole graph to perform related operations, these methods suffer poor scalability and stability. The idea of spatial-based method is to directly aggregate information from neighbors of each node. Atwood et al. \cite{atwood2016diffusion} regarded graph convolution as a diffusion process and assumed that information is transferred from a node to one of its neighboring nodes with a certain transition probability. Hamilton et al. \cite{hamilton2017inductive} utilized aggregator functions to aggregate the information from sampled neighbors. Benefiting from the wide application of attention mechanism which have achieved good performance in various fields \cite{vaswani2017attention}, Veli{\v{c}}kovi{\'c} et al. \cite{velivckovic2017graph} used the self-attention mechanism to aggregate node information based on neighbors' importance. The graph neural network models mentioned above are mostly only for homogeneous graphs, and cannot be directly applied to heterogeneous graphs.

\subsection{Heterogeneous Graph Embedding}
Heterogeneous graph embedding aims to embed heterogeneous graphs into low-dimensional vector spaces. Dong et al. \cite{dong2017metapath2vec} designed metapath-guided random walk to generate sequences as the input of skip-gram \cite{mikolov2013efficient} model to obtain the embedding of each node. Fu et al. \cite{fu2017hin2vec} performed multiple prediction training tasks to learn embedding of both nodes and metapaths in heterogeneous information networks. Shi et al. \cite{shi2018heterogeneous} designed a metapath-based random walk to generate the same type of node sequences and apply DeepWalk model to learn node representation. Chen et al. \cite{chen2018pme} decomposed the heterogeneous graph into several bipartite graphs, and then applied the LINE \cite{tang2015line} model to learn the embedding of each bipartite graph. Zhang et al. \cite{zhang2019shne} performed joint optimization of heterogeneous skip-gram and deep semantic encoding to capture semantic-aware representation of heterogeneous graph. The above researches are traditional graph representation learning models, which only considered the graph structure and ignored node attributes. There are also many models base on deep learning. Wang et al. \cite{wang2019heterogeneous} used the attention mechanism to aggregate information on metapath-based homogeneous graphs, and then use semantic attention to aggregate multiple metapath information. Zhang et al. \cite{zhang2019heterogeneous} jointly considered the heterogeneous content information and heterogeneous structural information. However, in their models, the information aggregation process was carried out only within the same type of nodes. Fu et al. \cite{fu2020magnn} proposed intra-metapath aggregation to aggregate all node information in each metapath instance, but they indiscriminately treated different types of nodes in the same way. 

The are some other heterogeneous graph neural network models that used different methodologies instead of metapath. Hong et al. \cite{hong2020attention} designed a type-aware attention layer, which embedded each node of the whole heterogeneous graph by jointing different types of adjacent nodes and associated linkages. Hu et al. \cite{hu2020heterogeneous} proposed a subgraph sampling method and designed graph transformer to directly aggregate features from heterogenous neighbors. Hu et al. \cite{hu2019adversarial} employed adversarial learning to capture rich semantics on heterogeneous graph, but the node attributes were not considered in this model. Since these studies used different methodologies, they can be regarded as parallel works to ours.

\section{Preliminaries and Problem Statement}

In this section, we give some important definitions related to heterogeneous graph that will be used in the paper. Table 1 summarizes frequently used notations in this paper.

\begin{table}[]
	\caption{Notations and  Definitions.}
	\label{Notations}
	\begin{tabular}{@{}cc@{}}
		\toprule
		Notations          & Definitions \\ \midrule
		$\mathbb{R}^{d}$ & $d$-dimensional Euclidean space \\
		$a$, ${\rm a}$, ${\rm A}$ & Scalar, vector, matrix \\
		$\mathcal{G}$ & A graph $\mathcal{G}=\left( \mathcal{V}, \mathcal{E} \right )$ \\
		$\mathcal{P}$ & The set of metapath $P$ in a graph \\
		$\mathcal{G}^{t}$ & The set of metapath-based subgraphs of $\mathcal{G}$ \\
		$\mathcal{G}^{ho}$ & The set of homogeneous subgraphs \\
		$\mathcal{G}^{he}$ & The set of heterogeneous subgraphs \\
		$G$ & A metapath-based subgraph $G \in \mathcal{G}^{t}$ \\
		$N_{v}^{G}$ & The set of neighbors of node $v$ in graph $G$ \\
		${\rm h}_{v}$ & Raw (attribute) feature vector of node $v$ \\
		${\rm z}_{v}$ & The finial embedding of node $v$ \\
		${\rm W}$, ${\rm b}$ & Weight matrix, bias \\
		$e_{vu}^{G}$ & Importance of node $u$ to $v$ in graph $G$ \\
		$w_{G_{i}}$ & Importance of graph $G_{i}$ \\
		$\alpha$, $\beta$ & Normalized attention weight \\
		$\sigma \left( \cdot \right)$ & Activation function \\
		$\left | \cdot \right |$ & The cardinality of a set \\
		$\left | \right |$ & Vector concatenation \\ \bottomrule
	\end{tabular}
\end{table}

\begin{definition}
	\textbf{Heterogeneous Graph} \cite{sun2013mining}. A heterogeneous graph is a graph $\mathcal{G}=\left(\mathcal{V}, \mathcal{E}\right)$ associates with a node type mapping function $\phi :\mathcal{V} \rightarrow \mathcal{A}$ and an edge type mapping function $\psi :\mathcal{E} \rightarrow \mathcal{R}$. $\mathcal{A}$ and $\mathcal{R}$ denote the predefined sets of node types and edge types, where $\left| \mathcal{A} \right| + \left| \mathcal{R} \right|>2$.
\end{definition}

\textbf{Example}. A heterogeneous graph composed of multiple types of nodes (Author (A), Paper (P), Venue (V)) and relations (authoring relation between authors and papers, publication relation between papers and venues) is shown in Figure \ref{Fig1}(a).

\begin{definition}
	\textbf{Metapath} \cite{sun2012mining}. A metapth $P$ is defined as a path in the form of $A_{1}\overset{R_{1}}{\rightarrow}A_{2}\overset{R_{2}}{\rightarrow}\cdots \overset{R_{l}}{\rightarrow} A_{l+1}$  (abbreviated as $A_{1}A_{2} \cdots A_{l+1}$ ), which describes a composite relation $R=R_{1}\circ R_{2}\circ \cdots \circ R_{l}$ between node types $A_{1}$ and $A_{l+1}$, where $\circ$ denotes the composition operator on relations. If the metapath $P$ is the same as the reverse metapath $P^{-1}$, then the metapath $P$ is symmetric.
\end{definition}

\textbf{Example}. Figure \ref{Fig1}(b) depicts four metapths: paper-author-paper (PAP), paper-venue-paper (PVP), paper-author (PA) and paper-venue (PV). Different metapaths represent different semantics. PAP means two papers are authored by the same author; PVP means two papers are published in the same venue; PA means the authoring relations between papers and authors; PV means publication relations between papers and venues.

\begin{figure*}[h]
	\centering
	\includegraphics[width=16cm]{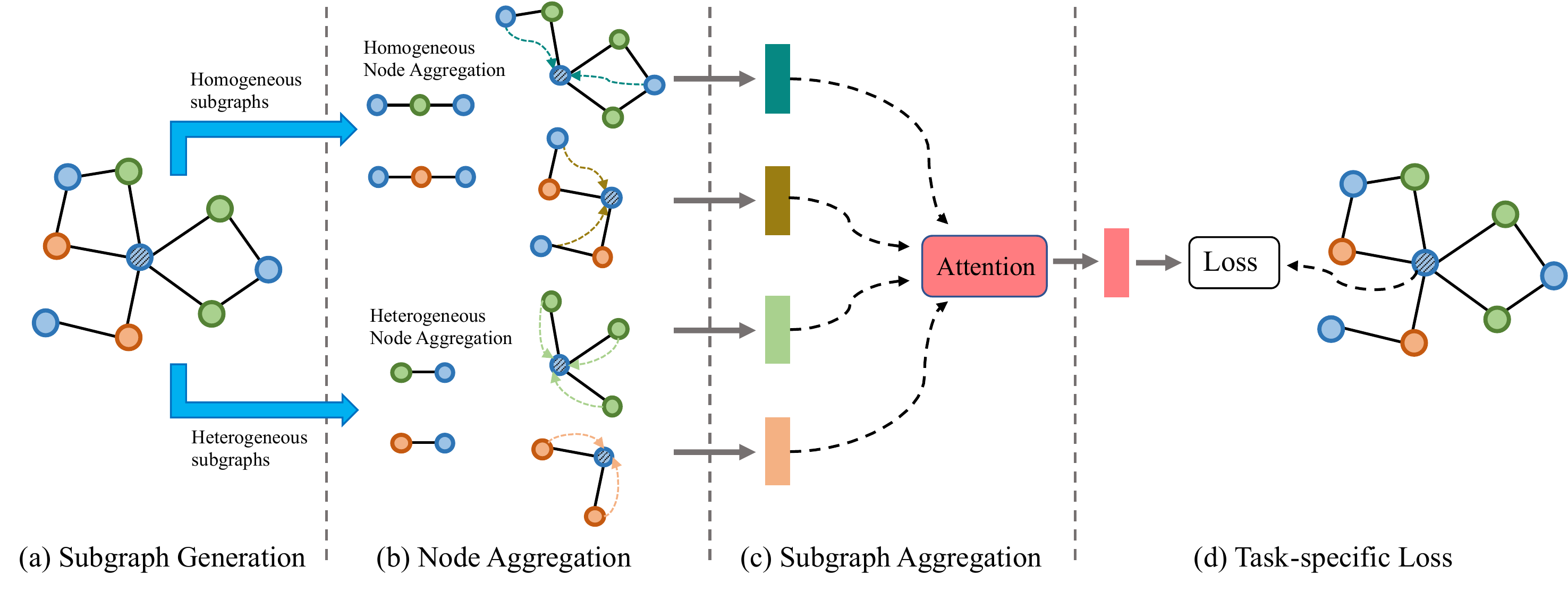}
	\caption{The overall architecture of HMSG. (a) Metapath-based subgraph generation; (b) Node aggregation within graphs; (c) Subgraph aggregation; (d) Loss function to be optimized.}
	\label{Fig2}
\end{figure*}

\begin{definition}
	\textbf{Metapath-based Neighbor}. Given a node $v$ and a metapath $P$ of a heterogeneous graph, the metapath-based neighbors $\mathcal{N}_{v}^{P}$ is defined as the set of nodes that connect to node $v$ via metapath $P$. Note that $\mathcal{N}_{v}^{P}$ includes $v$ itself if $P$ is symmetric.
\end{definition}

\textbf{Example}. Considering the metapath PAP in Figure \ref{Fig1}, the metapath-based neighbors of $p_1$ includes $p_1$ (itself), $p_2$; and $a_1$, $a_2$ are metapath-based neighbors of $p_1$ connected by metapath PA. In this paper, we unify the starting node in metapath as the target node.

\begin{definition}
	\label{definition4}
	\textbf{Metapath-based Subgraph}. Given a metapath $P$ of a heterogeneous graph $\mathcal{G}$, the metapath-based subgraph $\mathcal{G}^{P}$ is a graph constructed by all neighbor pairs based on metapath $P$ in graph $\mathcal{G}$. Note that $\mathcal{G}^{P}$ is a homogeneous subgraph if $P$ starts and ends with the same node type, otherwise it is a heterogeneous subgraph.
\end{definition}

\textbf{Example}. We give two metapath-based subgraphs in Figure \ref{Fig1}(c): A homogeneous subgraph generated by metapath PAP and a heterogeneous bipartite subgraph generated by metapath PA.

\textbf{Heterogeneous Graph Representation Learning Problem}: Given a heterogeneous graph $\mathcal{G}=\left( \mathcal{V},\mathcal{E} \right)$, the heterogeneous graph representation learning problem aims to learn $d$-dimensional node representations ${\rm z} \in \mathbb{R}^{\left | \mathcal{V} \right |\times d} \left ( d \ll \left | \mathcal{V} \right | \right )$ that are able to capture rich structural, semantic and attribute information.

\section{METHODOLOGY}
In this section, we formally present HMSG for heterogeneous graph embedding. HMSG consists of four parts: (1) node attribute transformation; (2) metapath-based subgraph generation; (3) node aggregation; and (4) subgraph aggregation. Figure \ref{Fig2} shows an overview about the framework of HMSG, and we will give detailed illustration in the following subsections.

\subsection{Node Attribute Transformation}
This part mainly aims to project attributes of different types of nodes into the same latent space, so that attribute information can be transferred among them. In a heterogeneous graph, different types of nodes are associated with different attributes, such as paper node with keywords, abstracts and author node with research fields. Those attributes are usually in represented in different feature spaces. Therefore, for each type of node, we design a type-specific linear transformation matrix to project their attributes into the same latent space. For a node $v \in \mathcal{V}_{A}$ of type $A$ in $\mathcal{A}$, we have 

\begin{equation}
	{{\rm h}}'_{v}={\rm W}_{A}\cdot {\rm h}_{v}^{A},
\end{equation}
where ${\rm h}_{v}^{A} \in \mathbb{R}^{d_{A}}$ is the original feature vector, and ${{\rm h}}'_{v} \in \mathbb{R}^{{d}'}$ is the projected feature vector of node $v$. ${\rm W}_{A} \in \mathbb{R}^{{d}' \times {d_{A}}}$ is the linear transformation matrix of node type $A$.

Through the transformation of attributes, the heterogeneity between different types of nodes is addressed, that facilitates the information transfer and aggregation between nodes in the graph.

\subsection{Metapath-based Subgraph Generation}
Different metapaths represent different semantics. According to the starting and ending node types of metapaths, we can divide metapaths into two categories for convenience:

\begin{equation}
	\mathcal{P}=\mathcal{P}^{t}, t \in \left \{ ho, he \right \},
\end{equation}
where $ho$ means the starting and ending nodes are of the same type in metapaths, otherwise is $he$.

In order to fully learn the information of each metapath, we generate the corresponding subgraphs according to Definition \ref{definition4} and then apply aggregation methods to each subgraph. Following the types $t$ of metapaths, the generated subgraphs can be divided into homogeneous subgraphs $\mathcal{G}^{ho}$ and heterogeneous subgraphs $\mathcal{G}^{he}$ (as shown in Figure \ref{Fig1}(c)):

\begin{equation}
	\mathcal{G}^{t}=\mathcal{G}^{ho} \cup \mathcal{G}^{he}.
\end{equation}

For a specific type of nodes, their connections to the neighbors in different subgraphs carry different semantic information, so each subgraph can be regarded as an interaction graph with specific semantic information. Since the subgraphs are independent, learning tasks can be carried out on each subgraph $G \in \mathcal{G}^{t}$ in parallel, which results in more efficient learning. Moreover, by learning from homogeneous and heterogeneous subgraphs independently, useful information can be retained as much as possible.

\subsection{Node Aggregation}
In this step, the information in each subgraph will be transmitted between nodes. For homogeneous graph learning, there are many excellent works such as GCN \cite{kipf2016semi}, GAT \cite{velivckovic2017graph} which can be used accordingly.

Each heterogeneous subgraph is in the form of a bipartite graph, because there are only two types of nodes in the subgraph and connections only exist between node pairs with different types. In bipartite graph aggregation, we are mainly concerned with the information of first-order heterogeneous neighbors, because the information of second-order neighbors, i.e., homogeneous neighbors, can be obtained from the homogeneous graph aggregation. Here we give three candidate aggregators inspired by GraphSAGE \cite{hamilton2017inductive}.

\textbf{Mean}. The mean operation averages the element-wise features of heterogeneous neighbor nodes as the features of target node:

\begin{equation}
	{\rm z}_{v}= {\rm MEAN} \left ( \left \{ {{\rm h}}'_{u}, \forall u \in N_{v} \right \} \right ),
\end{equation}
where $N_{v}$ is the neighbors of node $v$.

\textbf{Pooling}. Element-wise pooling operation aggregates information across heterogeneous neighbors in the following way: 

\begin{equation}
	{\rm z}_{v}= {\rm max}\left ( \left \{ \sigma \left ( {\rm W}_{\rm pool}{{\rm h}}'_{u}+{\rm b}_{\rm pool} \right ), \forall u \in N_{v} \right \} \right ),
\end{equation}
where $\sigma \left( \cdot \right)$ is the non-linearity activation function (ReLU in this paper), and $\rm max(\cdot)$ demotes the element-wise maximize operator which can be replaced by mean as well, ${\rm W}_{\rm pool} \in \mathbb{R}^{{d}' \times {d}'}$ and ${\rm b}_{\rm pool} \in \mathbb{R}^{{d}'}$ are learnable parameters.

\textbf{Attention}. The self-attention mechanism has been proven to be an effective information aggregator \cite{hamilton2017inductive}. Since different neighbors may have different influence on the target node, we use attention to learn the importance of each neighbor.

Next, we will use the self-attention mechanism to aggregate information from both homogeneous and heterogeneous graphs. Specifically, for a target node $v \in \mathcal{V}_{A}$, given a node pair $\left ( v, u \right )$ in $G \in \mathcal{G}_{A}^{t}$, which is generated by a metapath $P_{A}^{t}$ that starts from node type $A \in \mathcal{A}$, we adopt a graph attention layer to learn the importance $e_{vu}^{G}$ which measures how node $u$ would contribute to the target node $v$. In homogeneous subgraph the importance of node $u$ can be formulated as follows: 
\begin{equation}
	e_{vu}^{G}={\rm LeakyReLU} \left ( {\rm a}_{G}^{\rm T} \cdot \left [ {{\rm h}}'_{v} || {{\rm h}}'_{u} \right ] \right),
\end{equation}
while in heterogeneous subgraph the importance of node $u$ is:

\begin{equation}
	e_{vu}^{G}={\rm LeakyReLU} \left ( {\rm a}_{G}^{\rm T} \cdot {{\rm h}}'_{u}  \right), 
\end{equation}
where ${\rm a}_{G} \in \mathbb{R}^{2{d}'}$ in homogeneous graph and ${\rm a}_{G} \in \mathbb{R}^{{d}'}$  in heterogeneous graph are the parameterized attention vector for graph $G$ and $||$ denotes the concatenate operation.

Then we calculate $e_{vu}^{G}$ for nodes $u \in N_{v}^{G}$, where $N_{v}^{G}$ denotes the neighbors which are directly connected to node $v$ in $G$. We apply softmax function to obtain the normalized weight coefficient $\alpha_{vu}^{G}$: 

\begin{equation}
	\alpha_{vu}^{G}={\rm softmax} \left( e_{vu}^{G} \right ) = \frac{{\rm exp}\left( e_{vu}^{G} \right )}{\sum_{k \in N_{v}^{G}}{\rm exp}\left( e_{vk}^{G} \right )}.
\end{equation}

\begin{algorithm} 
	\caption{HMSG forward propagation}  
	\label{algorithm1} 
	\LinesNumbered  
	\KwIn{The heterogeneous graph $\mathcal{G}=\left ( \mathcal{V}, \mathcal{E} \right )$, \newline 
		node type $\mathcal{A}=\left \{ A_{1},A_{2},\cdots,A_{\left | \mathcal{A} \right |} \right \}$, \newline
		metapaths $\mathcal{P}=\mathcal{P}^{ho} \cup \mathcal{P}^{he}$, \newline
		subgraph type $\mathcal{T} = \left \{ ho, he \right \}$, \newline
		node features $\left \{ {\rm h}_{v}, \forall v \in \mathcal{V} \right \}$, \newline 
		the number of attention heads $K$
	}
	\KwOut{The node embeddings $\left \{ {\rm z}_{v}, \forall v \in \mathcal{V} \right \}$}  
	
	\For{node type $A \in \mathcal{A}$}{
		Type-specific transformation ${{\rm h}}'_{v}\leftarrow {\rm W}_{A}\cdot {\rm h}_{v}, \forall v \in \mathcal{V}_{A}$;
	}
	\For{node type $A \in \mathcal{A}$}{
		\For{$t \in \mathcal{T}$}{
			\For{metapath $P \in \mathcal{P}_{A}^{t}$}{
				Find the metapath-based subgraph $\mathcal{G}_{A}^{t}$; \newline
				\For{$v \in \mathcal{V}_{A}$}{
					Calculate the subgraph structure specific node embedding; \newline
					${\rm z}_{v}^{G} \leftarrow \mathop{||}\limits_{k=1}^K \sigma \left( \sum_{u \in N_{v}^{G}}\alpha_{vu}^{G} \cdot {\rm h}'_{u} \right )$;
				}
			}
		}
		Calculate the weight $\beta_{G}$ for each subgraph $ G \in \mathcal{G}_{A} $; \newline
		Fuse the embedding from different subgraphs \newline 
		${\rm z}_{v}^{G_{A}} \leftarrow \sum_{i=1}^{\left | \mathcal{G}_{A} \right |} \beta_{G_{i}} \cdot {\rm z}_{v}^{G_{i}}, \forall v \in \mathcal{V}_{A}$.
	}
\end{algorithm}

Finally, the embedding of node $v$ in subgraph $G$ can be aggregated by the neighbors' projected features with the corresponding coefficients as follows: 

\begin{equation}
	{\rm z}_{v}^{G}=\sigma \left( \sum_{u \in N_{v}^{G}} \alpha_{vu}^{G} \cdot {{\rm h}}'_{u}  \right ),
\end{equation}
where ${\rm z}_{v}^{G}$ is the output of node $v$ for subgraph $G$, and $\sigma \left ( \cdot \right )$ is the activation function.

In order to reduce the variance introduced by the heterogeneity of graphs and make the learning process more stable, we extended the self-attention to multiple heads. Specifically, we repeat $K$ independent attention mechanisms, and then concatenate the learned embeddings, resulting in the following formulation: 

\begin{equation}
	{\rm z}_{v}^{G}= \mathop{||}\limits_{k=1}^K \sigma \left( \sum_{u \in N_{v}^{G}}\alpha_{vu}^{G} \cdot {\rm h}'_{u} \right ).
\end{equation}

In summary, given $X$ metapath-based graphs $\mathcal{G}_{A}^{t}=\left \{ G_{1},\cdots, G_{X} \right\} $ with node type $A \in \mathcal{A}$ and projected node features ${\rm h}'$, $X$ groups of embeddings $\left \{ {\rm z}_{v}^{G_{1}}, \cdots, {\rm z}_{v}^{G_{X}} \right \} $ of the target node $v \in \mathcal{V}_{A}$ are generated, as shown in Figure \ref{Fig2}(b).

\subsection{Subgraph aggregation}
After node aggregation step, we obtain the embedding of each node in different subgraphs. In order to get more semantic information, attention mechanism is applied to assign different weights to different subgraphs according to their importance.

First, the embeddings we obtained from node aggregation are transformed through a nonlinear transformation. Then we sum up each subgraph $G_{i} \in \mathcal{G}_{A}$ by averaging the node embeddings for all nodes $v \in \mathcal{V}_{A}$, 

\begin{equation}
	w_{G_{i}}=\frac{1}{\left | \mathcal{V}_{A} \right |} \sum_{v \in \mathcal{V}_{A}}{\rm q}_{A}^{\rm T} \cdot {\rm tanh} \left( {\rm M}_{A} \cdot {\rm z}_{v}^{G_{i}} + {\rm b}_{A} \right ),
\end{equation}
where ${\rm q}_{A} \in \mathbb{R}^{d_{x}}$ is the parameterized attention vector for node type $A$, ${\rm M}_{A} \in \mathbb{R}^{d_{x}\times {d}'}$ and ${\rm b}_{A} \in\mathbb{R}^{d_{x}}$ are learnable parameters.

To make coefficients easily comparable across different subgraphs $G_{i}$, we normalize them using softmax function and then weighted summing all subgraphs: 

\begin{equation}
	\beta_{G_{i}}=\frac{{\rm exp} \left( w_{G_{i}} \right )}{\sum_{k=1}^{\left | \mathcal{G}_{A} \right |} {\rm exp} \left( w_{G_{k}} \right )},
\end{equation}

\begin{equation}
	{\rm z}_{v}^{G_{A}}=\sum_{i=1}^{\left | \mathcal{G}_{A} \right |} \beta_{G_{i}} \cdot {\rm z}_{v}^{G_{i}}.
\end{equation}

In this step, we get the final embeddings for all nodes $v \in \mathcal{V}_{A}$, Figure \ref{Fig2}(c) gives the illustration.

\subsection{Training}
Through the above sections, we obtain the node representations of each node, which can be used for different downstream tasks. Different loss function can be defined depend on specific tasks. We train HMSG in two major learning paradigms: semi-supervised learning and unsupervised learning.

For semi-supervised learning, there are only a small number of nodes in the graph with label information. We can minimize the cross entropy of labeled node data and apply back propagation and gradient descent methods to optimize the parameters of all nodes: 

\begin{equation}
	\mathcal{L}=-\sum_{v \in \mathcal{V}_L}y_{v}{\rm log}{y}'_v,
\end{equation}
where $\mathcal{V}_{L}$ is the set of labeled nodes, $y_v$ and ${y}'_{v}$ are the ground truth and predicted probability vector of node $v$, respectively.

For unsupervised learning, label information is invisible, we can optimize the model weights by minimizing the following reconstructive loss function through negative sampling \cite{mikolov2013distributed}: 

\begin{equation}
	\mathcal{L}=-\sum_{\left( v,u \right ) \in \mathcal{V}^+}{\rm log} \sigma \left({\rm h}_{v}^{\rm T}\cdot {\rm h}_{u} \right )-\sum_{\left( {v},{u}' \right ) \in \mathcal{V}^-}{\rm log} \sigma \left(- {\rm h}_{v}^{\rm T}\cdot {\rm h}_{{u}'} \right ), \label{Equ15}
\end{equation}
where $\sigma \left( \cdot \right)$ is the sigmoid function, $V^{+}$ is the set of connected (positive) node pairs, $V^{-}$ is the set of negative node pairs randomly sampled from all unconnected node pairs. The overall process of HMSG is shown in Algorithm \ref{algorithm1}.

\section{EXPERIMENTS}

\subsection{Datasets}
We use four commonly used public datasets to evaluate the performance of our proposed model and compare with state-of-the-art baselines. The detailed data description is summarized in Table \ref{Tab1}.

\begin{table}[]
	\caption{Statistics of datasets.}
	\label{Tab1}
	\begin{tabular}{@{}ccccc@{}}
		\toprule
		Datasets          & Node                                                                                                                   & Edge                                                                         & Metapath                                             \\ \midrule
		ACM              & \begin{tabular}[c]{@{}c@{}}\# Paper (P): 4,025\\ \# author (A): 7,167\\ \# Subject (S): 60 \\	\end{tabular}     & \begin{tabular}[c]{@{}cl@{}}\# P-A: 13,407 \\ \# P-S: 4,025\end{tabular} & \begin{tabular}[c]{@{}c@{}}PAP\\ PSP \\PA\\ PS \end{tabular} \\ \midrule
		IMDB             & \begin{tabular}[c]{@{}c@{}}\# movie (M): 4,181\\ \# actor (A): 5,257\\ \# director (D): 2,081 \\\end{tabular} & \begin{tabular}[c]{@{}c@{}}\# M-A: 12,537\\  \# M-D: 4,181\end{tabular} & \begin{tabular}[c]{@{}c@{}}MAM\\  MDM \\ MA\\ MD\end{tabular} \\ \midrule
		Amazon Review-I & \begin{tabular}[c]{@{}l@{}}\# user (U): 19,875\\ \# item (I): 24,340\end{tabular}                                & \# U-I: 180,047 & \begin{tabular}[c]{@{}c@{}} UIU \\ IUI \\ UI\\ IU\end{tabular} \\ \midrule
		Amazon Review-II& \begin{tabular}[c]{@{}l@{}}\# user (U): 16,437\\ \# item (I): 21,384\end{tabular}                                & \# U-I: 209,416 & \begin{tabular}[c]{@{}c@{}}UIU \\ IUI \\ UI\\ IU \end{tabular} \\ \bottomrule
	\end{tabular}
\end{table}

\textbf{ACM}$\footnote{http://dl.acm.org/}$. This is an academic network. Similar to HAN \cite{wang2019heterogeneous}, we extract a subset of ACM that contains 4025 papers (P), 7167 authors (A) and 60 subjects (S) according to the conferences (KDD, SIGMOD, VLDB, SIGCOMM, MobiCOMM) where papers published in, and divide the papers into three classes (Data Mining, Database, Wireless Communication). Each paper is described by a bag-of-words representation of terms. The metapaths we selected are \{PAP, PSP\} and \{PA, PS\}. For semi-supervised learning tasks, the paper nodes are divided into training, validation, and testing ratio of 1:1:8.

\textbf{IMDB}$\footnote{https://www.imdb.com/}$. This is an online database about movies and television programs. We extract a subset of IMDB that contains 4181 movies (M), 5257 actors (A), and 2081 directors (D). Each movie is labeled as one of three classes (Action, Comedy, Drama) based on their genre. Each movie is described by a bag-of-words representation of its plot keywords. The metapaths we selected are \{MAM, MDM\} and \{MA, MD\}. For semi-supervised learning tasks, the movie nodes are divided into training, validation, and testing ratio of 1:1:8.

\textbf{Amazon Review}$\footnote{http://jmcauley.ucsd.edu/data/amazon/index.html}$. This dataset is about product reviews from Amazon which is used for the link prediction task, and no label or feature is included in it. We extract two datasets, i.e., Review-I (Video Games category) with 6259 users and 20145 items, Review-II (Movies category) with 12453 users and 32395 items. The metapaths we selected are \{UI, IU\}. For unsupervised learning tasks, the user-item pairs are divided into training, validation, and testing ratio of two groups, i.e., 5:1:4 and 7:1:2 for both Review-I and Review-II. The node features of training dataset are pretrained by DeepWalk algorithm \cite{perozzi2014deepwalk}.

\begin{table*}[]
	\caption{Quantitative results (\%) on the node classification task.}
	\label{Tab2}
	\begin{tabular}{@{}ccccccccccc@{}}
		\toprule
		Datasets              & Metrics                   & Training & \multicolumn{1}{c}{DeepWalk} & \multicolumn{1}{c}{metapath2vec} & \multicolumn{1}{c}{HERec} & \multicolumn{1}{c}{GCN} & \multicolumn{1}{c}{GAT} & \multicolumn{1}{c}{HAN} &  \multicolumn{1}{c}{MAGNN} & HMSG   \\ \midrule
		\multirow{8}{*}{ACM}  & \multirow{4}{*}{Macro-F1} 
		&    20\%   & 79.80  & 87.26  & 77.16  & 89.92  & 88.55  & 89.16 & 90.49  & \textbf{91.36 } \\
		&  & 40\%   & 80.71  & 87.81  & 80.04  & 89.98  & 89.61  & 90.13 & 90.65  & \textbf{91.57 } \\
		&  & 60\%   & 80.83  & 87.69  & 80.90  & 89.91  & 89.92  & 90.45 & 90.60  & \textbf{91.72 } \\
		&  & 80\%   & 80.69  & 87.83  & 81.28  & 89.87  & 90.11  & 90.55 & 90.58  & \textbf{91.81 } \\  
		& \multirow{4}{*}{Micro-F1} 
		&    20\%   & 80.41  & 87.87  & 78.67  & 89.96  & 88.71  & 89.57 & 90.47  & \textbf{91.32 } \\
		&  & 40\%   & 81.58  & 88.41  & 81.46  & 90.02  & 89.72  & 90.20 & 90.66  & \textbf{91.54 } \\
		&  & 60\%   & 81.72  & 88.34  & 82.22  & 89.96  & 90.00  & 90.51 & 90.62  & \textbf{91.69 } \\
		&  & 80\%   & 81.58  & 88.47  & 82.48  & 89.87  & 90.15  & 90.54 & 90.57  & \textbf{91.73 } \\ \midrule
		\multirow{8}{*}{IMDB} & \multirow{4}{*}{Macro-F1} 
		&    20\%   & 50.03  & 40.22  & 46.14  & 53.83  & 55.69  & 58.03 & 60.83  & \textbf{60.89 } \\ 
		&  & 40\%   & 52.26  & 41.98  & 47.86  & 54.14  & 56.55  & 58.63 & 61.21  & \textbf{61.56 } \\ 
		&  & 60\%   & 52.71  & 43.06  & 49.15  & 54.19  & 56.98  & 59.05 & 61.16  & \textbf{61.85 } \\ 
		&  & 80\%   & 52.54  & 43.50  & 50.13  & 53.99  & 57.38  & 59.00 & 61.12  & \textbf{62.04 } \\
		& \multirow{4}{*}{Micro-F1} 
		&    20\%   & 51.45  & 41.12  & 47.25  & 54.20  & 55.82  & 58.25 & 61.15  & \textbf{61.16 } \\
		&  & 40\%   & 53.77  & 43.16  & 49.39  & 54.44  & 56.68  & 58.82 & 61.47  & \textbf{61.81 } \\
		&  & 60\%   & 54.30  & 44.29  & 50.76  & 54.49  & 57.12  & 59.24 & 61.40  & \textbf{62.12 } \\
		&  & 80\%   & 54.23  & 44.81  & 51.82  & 54.31  & 57.57  & 59.22 & 61.57  & \textbf{62.33 } \\ \bottomrule
	\end{tabular}
\end{table*}

\begin{table*}[]
	\caption{Quantitative  results (\%) on the node clustering task.}
	\label{Tab3}
	\begin{tabular}{@{}cccccccccc@{}}
		\toprule
		Datasets              & Metrics & \multicolumn{1}{c}{DeepWalk} & \multicolumn{1}{c}{metapath2vec} & \multicolumn{1}{c}{HERec} & \multicolumn{1}{c}{GCN} & \multicolumn{1}{c}{GAT} & \multicolumn{1}{c}{HAN} & \multicolumn{1}{c}{MAGNN} & HMSG   \\ \midrule
		\multirow{2}{*}{ACM}  & NMI   & 53.75  & 35.71   & 28.46  & 60.77  & 65.65  & 67.18  & 67.24  & \textbf{70.16 } \\
		& ARI  & 50.35  & 29.00  & 33.83  & 66.36  & 70.79  & 72.04  & 70.55 & \textbf{74.51 } \\ \midrule
		\multirow{2}{*}{IMDB} & NMI   & 5.66  & 0.31  & 0.54  & 8.34   & 11.03  & 12.13  & 13.64  & \textbf{14.52 } \\
		& ARI  & 6.65  & 0.10  & 0.13  & 7.52  & 11.01  & 12.08  & 13.58 & \textbf{14.77 } \\ \bottomrule
	\end{tabular}
\end{table*}

\subsection{Baselines}
We compare HMSG with other graph embedding methods to verify the performance of our proposed method. The detail description of baselines are as follows.

\begin{itemize}
	\item \textbf{Deepwalk} \cite{perozzi2014deepwalk}: This is a random walk-based homogeneous graph representation learning method. We ignore the heterogeneity of the graph and input the entire graph into the DeepWalk model.
	\item \textbf{metapath2vec} \cite{dong2017metapath2vec}: This is a heterogeneous graph representation learning method based on metapath guided random walk and skip-gram model. We test all the metapaths for metapath2vec and report the best performance.
	\item \textbf{HERec} \cite{shi2018heterogeneous}: This is a heterogeneous graph embedding method which designs a metapath-based random walk to generate the same type of node sequences and feed into DeepWalk model. We test all the metapaths for HERec and report the best performance.
	\item \textbf{GCN} \cite{kipf2016semi}: It is a homogeneous graph convolutional network which aggregates information from immediate neighbors. We test GCN on metapath-based homogeneous graphs and report the results from the best metapath on semi-supervised learning tasks. On unsupervised learning tasks, we input the entire graph by ignoring the heterogeneity of the graph.
	\item \textbf{GAT} \cite{velivckovic2017graph}: It is a homogeneous graph convolutional network which calculates the importance of different neighbors by attention mechanism. We test GAT on metapath-based homogeneous graphs and report the results from the best metapath on semi-supervised learning tasks and input the entire graph by ignoring the heterogeneity of the graph on unsupervised learning tasks.
	\item \textbf{HAN} \cite{wang2019heterogeneous}: It is a heterogeneous graph neural network which designs node attention for aggregating metapath-based homogeneous graph and utilize semantic attention to aggregate the information of multiple metapaths.
	\item \textbf{MAGNN} \cite{fu2020magnn}: It is a heterogeneous graph neural network which ultilizes intra-metapath aggregation and inter-metapath aggregation to aggregate node information.
\end{itemize}

For random-walk-based methods, including DeepWalk, metapath2vec and HERec, we set window size to 5, walk length to 100, walk per node to 40, and negative samples to 5. For GCN, GAT, HAN and HMSG, we use the same splits of training, validation, and testing sets. We train them for 1000 epochs and apply early stopping with patience of 30. We employ the Adam optimizer with the learning rate set to 0.005 and the weight decay (L2 penalty) set to 0.001. We set the dropout rate to 0.6. In the HMSG model, we use the self-attention mechanism to aggregate homogeneous and heterogeneous subgraphs, and other variants of HMSG are analyzed in the ablation study section. For GAT, HAN and HMSG, we set the number of attention heads to 8. For HAN, HMSG, we set the dimension of attention vector in subgraph aggregation to 128. For a fair comparison, we set the embedding dimension of all the algorithms to 64. Our model is implemented via the Deep Graph Library (DGL) \cite{wang2019deep} package of PyTorch frameworks.

\begin{table*}
	\caption{Quantitative  results (\%) on the link prediction task.}
	\label{Tab4}
	\begin{tabular}{@{}cccccccccc@{}}
		\toprule
		Datasets                                                                            & \multicolumn{1}{c}{Metrics} & \multicolumn{1}{c}{DeepWalk} & \multicolumn{1}{c}{metapath2vec} & \multicolumn{1}{c}{HERec} & \multicolumn{1}{c}{GCN} & \multicolumn{1}{c}{GAT} & \multicolumn{1}{c}{HAN} & \multicolumn{1}{c}{MAGNN} & HMSG   \\ \midrule
		\multirow{2}{*}{\begin{tabular}[c]{@{}c@{}} Review-I\\(7:1:2)\end{tabular}} 
		& AUC & 63.66 & 68.96 & 62.55  & 79.78  & 76.79  & 84.12  & 84.36 & \textbf{84.56} \\
		& AP  & 59.57  & 65.74  & 58.78  & 80.40  & 75.16 & 82.76 & 83.43 & \textbf{83.66} \\
		\multirow{2}{*}{\begin{tabular}[c]{@{}c@{}} Review-I\\(5:1:4)\end{tabular}} 
		& AUC  & 60.11   & 68.52   & 59.78 & 76.12 & 72.10 & 79.11  & 79.17  & \textbf{79.72} \\
		& AP  & 55.56  & 65.31  & 53.84   & 77.05 & 70.57  & 78.37  & 77.05 & \textbf{79.08} \\ \midrule
		\multirow{2}{*}{\begin{tabular}[c]{@{}c@{}} Review-II\\ (7:1:2)\end{tabular}} 
		& AUC & 60.97  & 58.64  & 58.69  & 73.77  & 78.93  & 81.05  & 78.45  & \textbf{81.70} \\
		& AP  & 57.37  & 56.83  & 55.37  & 73.67  & 78.65 & 80.54 &  78.12 & \textbf{81.63} \\
		\multirow{2}{*}{\begin{tabular}[c]{@{}c@{}} Review-II\\ (5:1:4)\end{tabular}} 
		& AUC & 56.67  & 57.32  & 56.04  & 70.76  & 72.18  & 74.24  & 73.31 & \textbf{76.33} \\ 
		& AP  & 54.58  & 55.47  & 54.12  & 70.83 & 72.67  & 74.09  & 72.60 & \textbf{76.58} \\ \bottomrule
	\end{tabular}
\end{table*}

\subsection{Node Classification}
We conduct experiments on ACM and IMDB datasets to compare the node classification performance of different models. We use the embedding of labeled nodes (paper nodes in ACM and movie nodes in IMDB) as the input of the SVM classifier, and divide the train/test ratio into different proportions. Similar to \cite{fu2020magnn}, only the nodes in the testing set of HMSG are used for the downstream training and evaluation of SVM. This strategy is also used for node clustering and link prediction tasks. In order to eliminate the variance caused by the unbalanced data division, we repeat each experiment for 10 times and report the average $Macro\text{-}F1$ and $Micro\text{-}F1$ as evaluation metrics. The node classification results are shown in Table \ref{Tab2}.

It can be seen from the Table \ref{Tab2} that, compared to the state-of-the-art models HAN and MAGNN, HMSG achieves the best performance on all experimental groups, showing the significant advantage of fusing the information from both homogeneous and heterogeneous subgraphs. In addition, the performance of metapath2vec method which considers the heterogeneous structure outperforms other random walk-based methods, but it is still not as good as GCN and GAT, which are based on graph convolution model.

\subsection{Node Clustering}
We conduct clustering experiments on ACM and IMDB datasets to evaluate the quality of embeddings learned by HMSG. For the clustering experiment, similar to node classification, we take the embeddings of labeled nodes in the testing set as the input of K-Means model, and use NMI and ARI metrics to evaluate the performance. Similarly, the results are average of 10 executions. 

The results are shown in Table \ref{Tab3}, from which we see that HMSG significantly outperforms all baselines, showing its strong capability of learning effective node representations. Besides, there is a significant performance gap between graph convolution-based methods (GCN, GAT, HAN, MAGNN and HMSG) and random walk-based methods, which again shows the power of graph convolutional model.

\subsection{Link Prediction}
Link prediction is used to test our algorithm's effectiveness in unsupervised learning task. We use two Amazon Review datasets Review-I and Review-II to evaluate the performance in link prediction task. We regard connected user-item pairs as positive node pairs and all unconnected links as negative node pairs. In the training phase, we randomly select the same number of positive and negative node pairs to training, validation and testing, and use formula \eqref{Equ15} to optimize the model. 

Then, we use the dot product of user and item embeddings obtained from the model to calculate the link probability between two nodes: $p_{ui}=\sigma \left( {\rm z}_{u}^{\rm T}\cdot {\rm z}_{i} \right)$, and $\sigma \left( \cdot \right )$ is the sigmoid function. We conduct 10 repeated experiments and report the average AUC and AP metrics on the testing set. The results of link prediction task are presented in Table \ref{Tab4}.

As shown in Table \ref{Tab4}, HMSG still performs the best on all the dataset groups evaluated by all the metrics. Again, graph convolution-based methods perform much better than random-walk-based methods. At the same time, we can see that heterogeneous graph models such as HAN and MAGNN perform better than homogeneous neural networks GCN and GAT in most of the time.

In sum, the above results validate the effectiveness of our HMSG model. In the following subsection, we will conduct ablation study to compare different variants of HMSG.

\begin{table*}
	\caption{Quantitative results (\%) for ablation study.}
	\label{Tab5}
	\begin{tabular}{lcccccccccccc}
		\toprule
		\multicolumn{1}{c}{\multirow{2}{*}{Variants}} & \multicolumn{4}{c}{ACM} & \multicolumn{4}{c}{IMDB} & \multicolumn{2}{c}{Review-I} & \multicolumn{2}{c}{Review-II} \\ \cline{2-13} 
		\multicolumn{1}{c}{} & Macro-F1 & Micro-F1 & NMI & ARI & Macro-F1 & Micro-F1 & NMI & ARI & AUC & AP & AUC & AP \\ \midrule 
		
		${\rm HMSG_{ho}}$ & 91.30 & 91.25 & 69.85 & 74.22 & 61.31 & 61.57 & \textbf{14.79} & \textbf{15.42} & 78.32 & 76.07 & 76.28 & 75.86 \\ 
		${\rm HMSG_{mean}}$ & 91.42 & 91.38 & 70.20 & 74.75 &61.40 & 61.72 & 14.36 & 14.41 & 78.94 & 77.49 & 76.24 & 76.48 \\ 
		${\rm HMSG_{pool}}$ & 91.48 & 91.42 & \textbf{70.78} & \textbf{75.13} & 61.43 & 61.73 & 14.29 & 14.22 & 79.06 & 77.41 & 74.24 & 73.70 \\ 
		${\rm HMSG_{atten}}$ & \textbf{91.62} & \textbf{91.57} & 70.16 & 74.51 & \textbf{61.59} & \textbf{61.86} & 14.52 & 14.77 & \textbf{82.14} & \textbf{81.37} & \textbf{78.96} & \textbf{79.05} \\  \bottomrule
	\end{tabular}
\end{table*}

\subsection{Ablation Study}
To validate the effectiveness of different components in our model, we conduct experiments on different variants of HMSG. Here ${\rm HMSG_{\rm ho}}$ means only homogeneous subgraphs are used in the model; ${\rm HMSG_{\rm mean}}$ means we use an average strategy in heterogeneous subgraph aggregation; ${\rm HMSG_{\rm pool}}$ means max pooling strategy in heterogeneous subgraph aggregation; ${\rm HMSG_{\rm atten}}$ means self-attention mechanism are used in heterogeneous subgraph aggregation. We present the average results with different training ratios in Table \ref{Tab5}.

As can be seen, except for the clustering task on IMDB where the ${\rm HMSG_{\rm ho}}$ model (which uses homogeneous subgraphs only) performs the best, on all the other experimental groups, HMSG models considering both homogeneous and heterogeneous subgraphs give the best results, among which ${\rm HMSG_{\rm atten}}$ model exhibits the best overall performance, which again validates the effectiveness and necessity of our model design.

\section{CONCLUSION}

In this paper we proposed a new heterogeneous graph neural network model HMSG, which learned graph representations from multiple homogeneous and heterogeneous subgraphs generated through various metapaths. Compared to existing models, HMSG is able to comprehensively capture structural, semantic and attribute information from both homogeneous and heterogeneous neighbors. Extensive experiments on multiple tasks and datasets showed that HMSG significantly outperformed the state-of-the-art baselines. 

For the future work, we will consider extending HMSG so that it can support dynamic heterogeneous graphs. We will also try other subgraph generating and aggregating methods in the HMSG model in the future.


\bibliographystyle{abbrv}
\bibliography{Reference}
\end{document}